\documentclass[11pt,a4paper]{article}
\usepackage[hyperref]{emnlp2020}
\usepackage{times}
\usepackage{latexsym}
\usepackage{booktabs}

\usepackage{amsmath}
\usepackage{graphicx}
\usepackage{bm}
\usepackage{multirow}
\usepackage{subcaption} 
\usepackage{comment}
\usepackage{xspace}

\DeclareMathOperator*{\En}{\text{En}}
\DeclareMathOperator*{\Fr}{\text{Fr}}
\DeclareMathOperator*{\De}{\text{De}}
\DeclareMathOperator*{\GSTE}{\text{GSTE}}
\DeclareMathOperator*{\INT}{\text{INT}}
\DeclareMathOperator*{\SU}{\text{CE}}
\DeclareMathOperator*{\S2P}{\text{S2P}}

\DeclareMathOperator*{\TEACHER}{\text{TEACHER}}
\DeclareMathOperator*{\STUDENT}{\text{STUDENT}}

\newcommand{\algo}{SSIL\xspace}
\newcommand{\longalgo}{Supervised Seeded Iterated Learning\xspace}
\usepackage{microtype}

\aclfinalcopy 
\def\aclpaperid{2583} 

\title{Supervised Seeded Iterated Learning for Interactive Language Learning}

\author{
  Yuchen Lu \\
  Mila \\
  University of Montreal \\\And

  Soumye Singhal \\
  Mila \\
  University of Montreal \\
 
  \And
  Florian Strub   \\
  DeepMind \\
  
  \AND
  Olivier Pietquin \\
  Google Research \\
  Brain Team \\
  \And

  Aaron Courville  \\
  Mila\\
  University of Montreal, CIFAR \\
 }

\date{}

\begin{document}
\maketitle
\begin{abstract}
Language drift has been one of the major obstacles to train language models through interaction.
When word-based conversational agents are trained towards completing a task, they tend to invent their language rather than leveraging natural language.
In recent literature, two general methods partially counter this phenomenon: Supervised Selfplay (S2P) and Seeded Iterated Learning (SIL). 
While S2P jointly trains interactive and supervised losses to counter the drift, SIL changes the training dynamics to prevent language drift from occurring.
In this paper, we first highlight their respective weaknesses, i.e., late-stage training collapses and higher negative likelihood when evaluated on human corpus.
Given these observations, we introduce \longalgo~(\algo) to combine both methods to minimize their respective weaknesses. 
We then show the effectiveness of \algo in the language-drift translation game.   
\end{abstract}

\section{Introduction}
Since the early days of NLP~\cite{winograd1971procedure}, conversational agents have been designed to interact with humans through language to solve diverse tasks, e.g., remote instructions~\cite{thomason2015learning} or booking assistants \cite{bordes2016learning,elasri2017frames}. In this goal-oriented dialogue setting, the conversational agents are often designed to compose with predefined language utterances~\cite{lemon2007machine,williams2014dialog,young2013pomdp}. Even if such approaches are efficient, they also tend to narrow down the agent's language diversity. 
To remove this restriction, recent work has been exploring interactive word-based training.
In this setting, the agents are generally trained through a two-stage process~\cite{wei2018airdialogue, de2017guesswhat, shah2018bootstrapping, li2016dialogue, das2017learning}: Firstly, the agent is pretrained on a human-labeled corpus through supervised learning to generate grammatically reasonable sentences. Secondly, the agent is finetuned to maximize the task-completion score by interacting with a user. Due to sample-complexity and reproducibility issues, the user is generally replaced by a game simulator that may evolve with the conversational agent. Unfortunately, this pairing may lead to the \emph{language drift} phenomenon, where the conversational agents gradually co-adapt, and drift away from the pretrained natural language. The model thus becomes unfit to interact with humans~\cite{chattopadhyay2017evaluating, zhu2017interactive, lazaridou2020multi}.

While domain-specific methods exist to counter language drift~\cite{lee2019countering,li2016deep}, a simple task-agnostic method consists of combining interactive and supervised training losses on a pretraining corpus~\cite{wei2018airdialogue, lazaridou2016multi}, which was later formalized as Supervised SelfPlay (S2P) ~\cite{lowe2020on}. 

\begin{figure*}[ht]
    \begin{center}
    \includegraphics[width=0.9\linewidth]{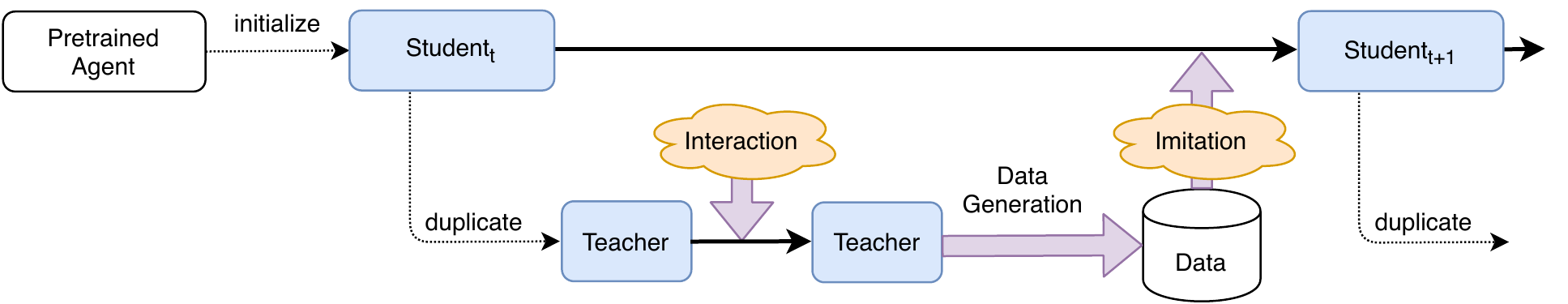}
    \vskip -0.6em
    \caption{SIL~\cite{lu2020countering}. A student agent is iteratively refined using newly generated data from a teacher agent. At each iteration, a teacher agent is created on top of the student before being finetuned by interaction, e.g. maximizing a task completion-score. Teacher generates a  dataset with greedy sampling and students imitate those samples. The interaction step involves interaction with another language agent. 
    }
    \label{fig:sil}
    \end{center}
    \vskip -1em
\end{figure*}

Inspired by language evolution and cultural transmission~\cite{kirby2001spontaneous, kirby2014iterated},
recent work proposes Seeded Iterated Learning (SIL)~\cite{lu2020countering} as another task-agnostic method to counter language drift.
SIL modifies the training dynamics by iteratively refining a pretrained student agent by imitating interactive agents, as illustrated in Figure~\ref{fig:sil}. At each iteration, a teacher agent is created by duplicating the student agent, which is then finetuned towards task completion. A new dataset is then generated by greedily sampling the teacher, and those samples are used to refine the student through supervised learning. The authors empirically show that this iterated learning procedure induces an inductive learning bias that successfully maintains the language grounding while improving task-completion. 

As a first contribution, we further examine the performance of these two methods in the setting of a translation game~\cite{lee2019countering}. 
We show that S2P is unable to maintain a high grounding score and experiences a late-stage collapse, while SIL has a higher negative likelihood when evaluated on human corpus.

We propose to combine SIL with S2P by applying an S2P loss in the interactive stage of SIL. We show that the resulting \emph{Supervised Seeded Iterated Learning} (\algo) algorithm manages to get the best of both algorithms in the translation game.
Finally, we observe that the late-stage collapse of S2P is correlated with conflicting gradients before showing that \algo empirically reduces this gradient discrepancy.

\section{Preventing Language Drift}

We describe here our interactive training setup before introducing different approaches to prevent language drift. In this setting, we have a set of collaborative agents that interact through language to solve a task. To begin, we train the agents to generate natural language in a word-by-word fashion. Then we finetune the agents to optimize a task completion score through interaction, i.e., learning to perform the task better. Our goal is to prevent the language drift in this second stage.

\subsection{Initializing the Conversational Agents}
For a language agent $f$ parameterized by $\bm{\theta}$, and a sequence of generated words $\bm{w}_{1:i} = [w_{j}]_{j=1}^i$ and an arbitrary context $\bm{c}$, the probability of the next word $w_i$ is $p(w_{i+1}|\bm{w}_{1:i}, \bm{c})=f_{\bm{\theta}}(\bm{w}_{1:i}, \bm{c})$
We pretrain the language model to generate meaningful sentences by minimizing the cross-entropy loss $\mathcal{L}^{\SU}_{pretrain}$ where the word sequences are sampled from a language corpus $D_{pretrain}$. 
Note that this language corpus may either be task-related or generic. Its role is to get our conversational agents a reasonable initialization. 

\subsection{Supervised Selfplay (S2P)}
A common way to finetune the language agents while preventing language drift is to replay the pretraining data during the interaction stage. In S2P the training loss encourages both maximizing task-completion while remaining close to the initial language distribution. Formally, 
\begin{equation}
\mathcal{L}^{\S2P} = \mathcal{L}^{\INT} + \alpha \mathcal{L}^{\SU}_{pretrain}
\end{equation}
where $\mathcal{L}^{\INT}$ is a differentiable interactive loss maximizing task completion, e.g. reinforcement learning with policy gradients~\cite{sutton2000policy}, Gumbel Straight-through Estimator (STE)~\cite{jang2017categorical} etc., $\mathcal{L}^{\SU}_{pretrain}$ is a cross-entropy loss over the pretraining samples. $\alpha$ is a positive scalar which balances the two losses.

\subsection{Seeded Iterated Learning (SIL)}
Seeded Iterated Learning (SIL) iteratively refines a pretrained \emph{student} model by using data generated from newly trained \emph{teacher} agents~\cite{lu2020countering}. As illustrated in Figure~\ref{fig:sil}, the student agent is initialized with the pretrained model. At each iteration, a new teacher agent is generated by duplicating the student parameters. It is tuned to maximize the task-completion score by optimizing the interactive loss $\mathcal{L}^{\TEACHER} = \mathcal{L}^{\INT}$
In a second step, we sample from the teacher to generate new training data $D_{teacher}$, and we refine the student by minimizing the cross-entropy loss
$\mathcal{L}^{\STUDENT} = \mathcal{L}^{\SU}_{teacher}$ where sequence of words are sampled from $D_{teacher}$. This imitation learning stage can induce an information bottleneck, encouraging the student to learn a well-formatted language rather than drifted components.

\subsection{\algo: Combining SIL and S2P}

S2P and SIL have two core differences: first, SIL never re-uses human pretraining data.
As observed in Section~\ref{sec:weaknesses}, this design choice reduces the language modeling ability of SIL-trained agents, with a higher negative likelihood when evaluated on human corpus. 
Second, S2P agents merge interactive and supervised losses, whereas SIL's student never experiences an interactive loss. As analyzed in Section~\ref{sec:gradient}, the S2P multi-task loss induces conflicting gradients, which may trigger language drift. 
In this paper, we propose to combine these two approaches and demonstrate that the combination effectively minimizes their respective weaknesses. To be specific, we apply the S2P loss over the SIL teacher agent, which entails $\mathcal{L}^{\TEACHER} = \mathcal{L}^{\INT} + \alpha \mathcal{L}^{\SU}_{pretrain}$. We call the resulting algorithm, \longalgo~(\algo). In \algo, teachers can generate data that is close to the human distribution due to the S2P loss, while students are updated with a consistent supervised loss to avoid the potential weakness of multi-task optimization. In addition, \algo still maintains the inductive learning bias of SIL.
We list all these methods in Table~\ref{tab:sil_s2p} for easy comparison.
We also experiment with other ways of combining SIL and S2P by mixing the pretraining data with teacher data during the imitation learning stage. We call this method \emph{MixData}. We show the results of this approach in Appendix~\ref{sec:mixdata}. We find that this approach is very sensitive to the mixing ratio of these two kinds of data, and the best configuration is still not as good as \algo.

\begin{table}[t]
\centering
{
\small
\begin{tabular}{ll}
\toprule
\emph{Finetuning Methods} & \emph{Training Losses}                                    \\  \midrule
Gumbel          & $\mathcal{L}^{\INT} $                   \\ \midrule
S2P              & $\mathcal{L}^{\INT} + \alpha \mathcal{L}^{\SU}_{pretrain}$   \\ \midrule
SIL (teacher)    & $\mathcal{L}^{\INT}$\\ 
SIL (student)    & $\mathcal{L}^{\SU}_{teacher}$ \\ \midrule
\algo (teacher) & $\mathcal{L}^{\INT} + \alpha \mathcal{L}^{\SU}_{pretrain}$\\ 
\algo (student) & $\mathcal{L}^{\SU}_{teacher}$  \\ \bottomrule
\end{tabular}
}
\caption{Finetuning with respective training objective.}
\label{tab:sil_s2p}
\vskip -1em
\end{table}

\section{Experimental Setting}
\label{sec:settings}
\subsection{Translation Game}
We replicate the translation game setting from ~\cite{lee2019countering} as it was designed to study language drift. First, a \emph{sender} agent translates French to English (Fr-En), while a \emph{receiver} agent translates English to German (En-De). The sender and receiver are then trained together to translate French to German with English as a pivot language.
For each French sentence, we sample English from the sender, send it to the receiver, and sample German from the receiver.

The task score is defined as the BLEU score between generated German translation and the ground truth (\emph{BLEU De})~\cite{papineni2002bleu}. 
The goal is to improve the task score without losing the language structure of the intermediate English language.
\subsection{Training Details}
The sender and the receiver are pretrained on the IWSLT dataset~\cite{cettolo2012wit3} which contains $(\Fr, \En)$ and $(\En, \De)$ translation pairs. We then use the Multi30k dataset~\citep{elliott2016multi30k} to build the finetuning dataset with $(\Fr, \De)$ pairs. 
As IWSLT is a generic translation dataset and Multi30k only contains visually grounded translated captions, we also call IWSLT task-agnostic while Multi30K task-related.
We use the cross-entropy loss of German as the interactive training objective, which is differentiable w.r.t. the receiver. For the sender, we use Gumbel Softmax straight-through estimator to make the training objective also differentiable w.r.t. the sender, as in~\citet{lu2020countering}. 

Implementation details are in Appendix~\ref{sec:translation_game} 
\subsection{Metrics for Grounding Scores}
In practice, there are different kinds of language drift~\cite{lazaridou2020multi} (e.g. syntactic drift and semantic drift). We thus have multiple metrics to consider when evaluating language drift.
We first compute English BLEU score (\emph{BLEU En}) comparing the generated English translation with the ground truth human translation. We include the negative log-likelihood (\emph{NLL}) of the generated En translation under a pretrained language model as a measure of syntactic correctness. 
In line with \cite{lu2020countering} , we also report results using another language metric: the negative log-likelihood of human translations (\emph{RealNLL}) given a finetuned Fr-En model. We feed the finetuned sender with human task-data to estimate the model's log likelihood. The lower is this score, the more likely the model would generate such human-like language.

\section{Experiments}
\begin{figure*}[th!]
  \begin{subfigure}{0.47\columnwidth}
  \includegraphics[width=\textwidth]{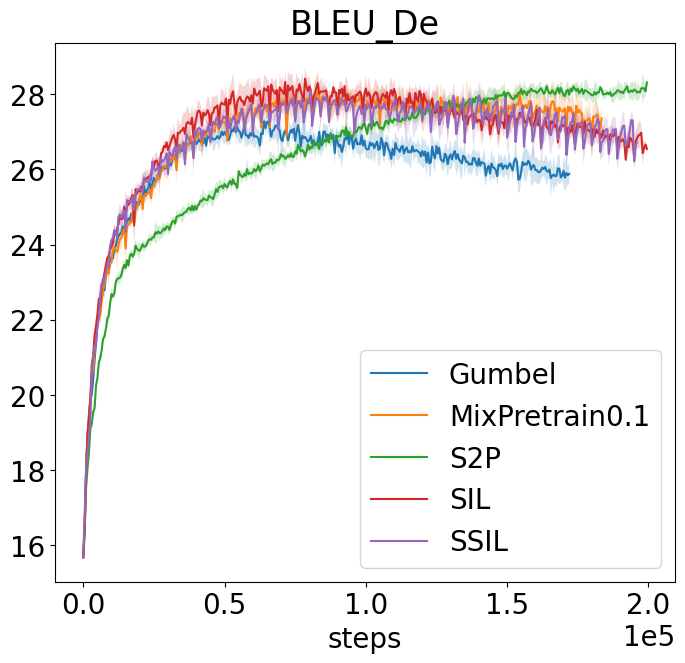}
  \caption{BLEU De (Task Score)}
  \end{subfigure}
  \hfill
  \begin{subfigure}{0.47\columnwidth}
  \includegraphics[width=\textwidth]{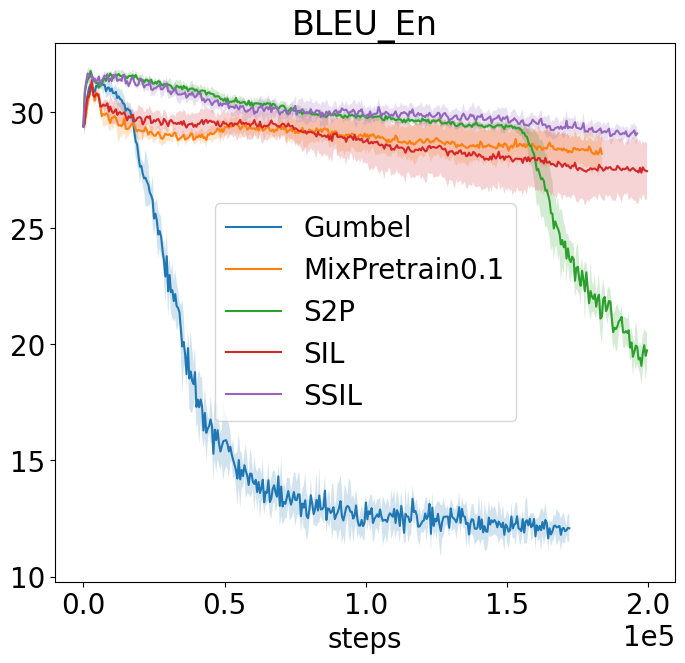}
  \caption{BLEU En} 
  \end{subfigure} 
  \hfill  
  \begin{subfigure}{0.47\columnwidth} 
  \includegraphics[width=\textwidth]{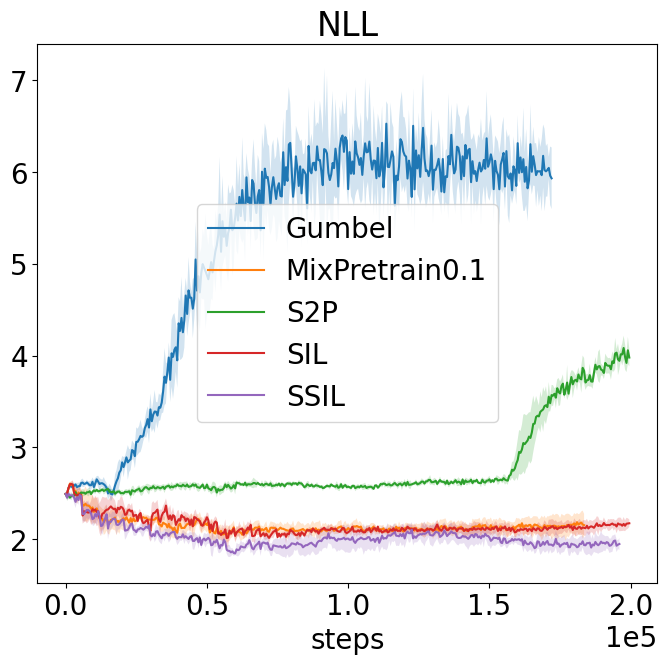} 
  \caption{NLL} 
  \end{subfigure}  
  \hfill 
  \begin{subfigure}{0.47\columnwidth} 
  \includegraphics[width=\textwidth]{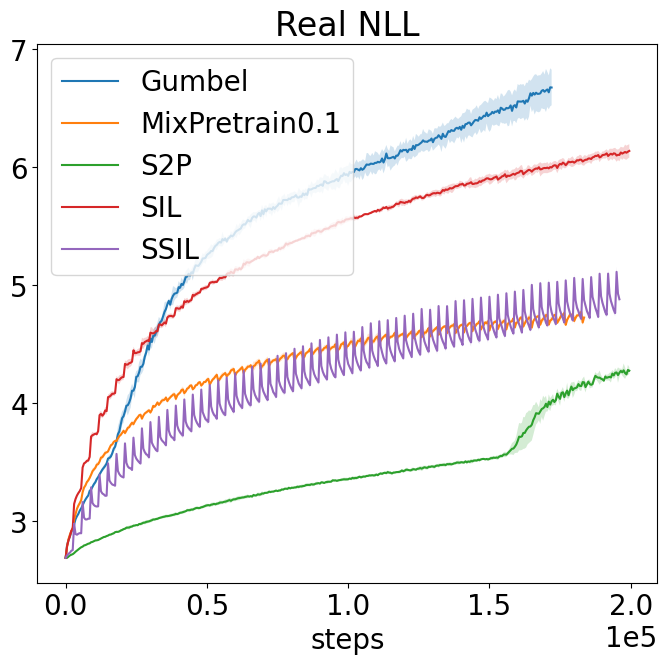} 
  \caption{RealNLL} 
  \end{subfigure}
  \caption{Task and language metrics for Vanilla Gumbel, SIL, S2P, and \algo in the translation game average over 5 seeds. We also show the results of mixing pretraining data in the teacher dataset (Section~\ref{sec:mixdata}). The plots are averaged over 5 seeds with shaded area as standard deviation. Although SIL and S2P both counter language drift, S2P suffers from late collapse, and SIL has a high \emph{RealNLL}, suggesting that its output may not correlate well with human sentences. }
  \label{fig:problem}
  \vskip -0.5em
\end{figure*}

\begin{figure}[t]

  \begin{subfigure}{0.47\columnwidth}
    \includegraphics[width=\textwidth]{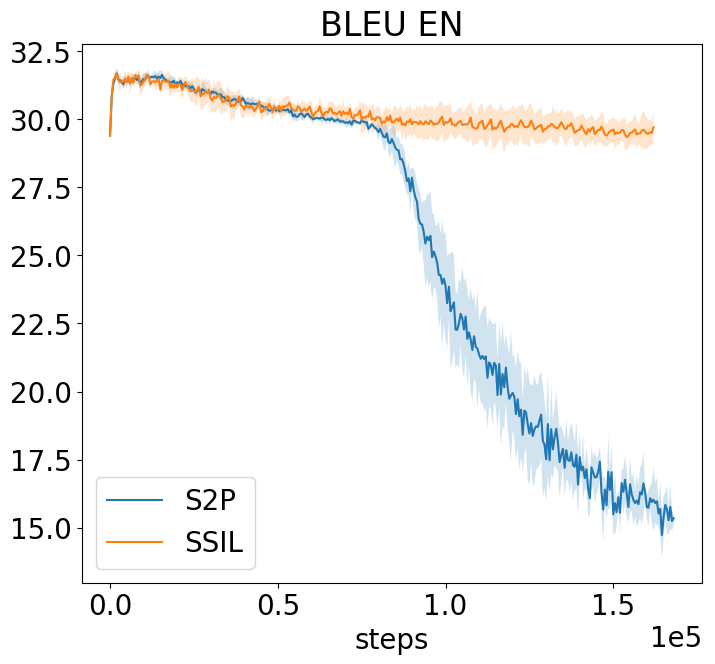}
    \caption{Bleu En}
  \end{subfigure}
  \hfill
  \begin{subfigure}{0.47\columnwidth}
    \includegraphics[width=\textwidth]{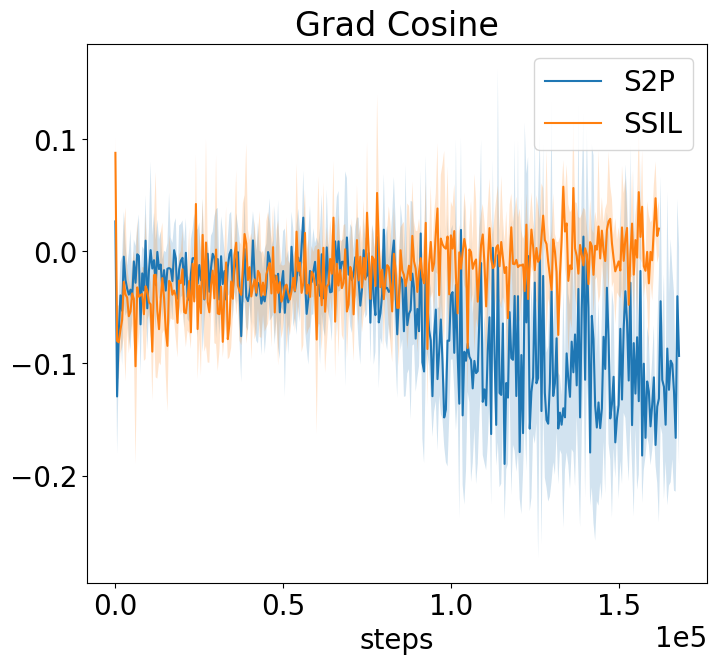}
    \caption{Cosine Similarity}
  \end{subfigure}
  \caption{Cosine similarity between the gradients issued from $\mathcal{L}^{\INT}$ and $\mathcal{L}^{\SU}_{pretrain}$. The collapse of the BLEU En matches the negative cosine similarity.We here set $\alpha=0.5$ but similar values yield identical behavior as shown in Figure~\ref{fig:grad_cosine_0_7} in Appendix.
  }
  \label{fig:grad_cosine}
  \end{figure}

\subsection{S2P and SIL Weaknesses}
\label{sec:weaknesses}

We report the task and grounding scores of vanilla Gumbel, S2P, SIL, and \algo in Figure~\ref{fig:problem}. The respective best hyper-parameters can be found in the appendix. As reported by~\citet{lu2020countering}, vanilla Gumbel successfully improves the task score \emph{BLEU De}, but the \emph{BLEU En} score as well as other grounding metric collapses, indicating a language drift during the training. Both S2P and SIL manage to increase \emph{BLEU De} while maintaining a higher \emph{BLEU En} score, countering language drift. However, S2P has a sudden (and reproducible) late-stage collapse, unable to maintain the grounding score beyond 150k steps. On the other hand, SIL has a much higher RealNLL than S2P, suggesting that SIL has a worse ability to model human data.

\algo seems to get the best of the two worlds. It has a similar task score \emph{BLEU De} as S2P and SIL, while it avoids the late-stage collapse. It ends up with the highest \emph{BLEU En}, and it improves the RealNLL over SIL, though still not as good as S2P. Also, it achieves even better NLL, suggesting that its outputs are favoured by the pretrained language model.

\subsection{Mixing Teacher and Human Data}
\label{sec:mixdata}
We also explore whether injecting pretraining data into the teacher dataset may be a valid substitute for the S2P loss. We add a subset of the pretraining data in the teacher dataset before refining the student, and we report the results in Figure~\ref{fig:problem} and~\ref{fig:mixdata}. Unfortunately, such an approach was quite unstable, and it requires heavy hyper-parameters tuning to match \algo scores. As explained in ~\cite{kirby2001spontaneous}, iterated learning rely on inductive learning to remove language irregularities during the imitation step. Thus, mixing two language distributions may disrupt this imitation stage.

\subsection{Why S2P collapses?}
\label{sec:gradient}
We investigate the potential cause of S2P late-stage collapse and how \algo may resolve it. 
We firstly hope to solve this by increasing the supervised loss weight $\alpha$.
However, we find that a larger $\alpha$ only delays the eventual collapse as well as decreases the task score, as shown in Figure~\ref{fig:s2p} in Appendix~\ref{sec:s2p}.

We further hypothesize that this late-stage collapse can be caused by the distribution mismatch between the pretraining data (IWSLT) and the task-related data (Multi30K), exemplified by their word frequencies difference.
A mismatch between the two losses could lead to conflicting gradients, which could, in turn, make training unstable. In Figure~\ref{fig:grad_cosine}, we display the cosine similarity of the sender gradients issued by the interactive and supervised losses $cos(\nabla_{sender} \mathcal{L}^{\INT}$, $\nabla_{sender}\mathcal{L}^{\SU}_{pretrain})$ for both S2P and \algo for $\alpha=0.5$ during training. Early in S2P training, we observe that the two gradients remain orthogonal on average, with the cosine oscillating around zero. 
Then, at the same point where the S2P \emph{Bleu En} collapses, the cosine of the gradients starts trending negative, indicating that the gradients are pointing in opposite directions. However, \algo does not have this trend, and the \emph{BLEU En} does not collapse. 
Although the exact mechanism of how conflicting gradients trigger the language drift is unclear, current results favor our hypothesis and suggest that language drift could result from standard multi-task optimization issues \cite{yu2020gradient, parisotto2015actor, sener2018multi} for S2P-like methods.

\paragraph{Conclusion}
We investigate two general methods to counter language drift: S2P and SIL. S2P experiences a late-stage collapse on the grounding score, whereas SIL has a higher negative likelihood on human corpus. We introduce \algo to combine these two methods effectively. We further show the correlation between S2P late-stage collapse and conflicting gradients.

\paragraph{Acknowledgement}
We thank Compute Canada (www.computecanada.ca) for providing the computational resources. We thank Miruna Pislar and Angeliki Lazaridou for their helpful discussions.

{
\small
\bibliographystyle{acl_natbib}
\bibliography{anthology, emnlp2020}
}
\newpage
\appendix

\onecolumn

\section{Explicit losses in the Translation Game}
\paragraph{S2P}
Let $\mathcal{L}^{\GSTE}(Fr, De)$ be the loss of Gumbel STE, when two agents is fed with $Fr$ and the ground truth German translation $De$. Let $\mathcal{L}^{\SU}(X, Y)$ to be the supervised training loss with source $X$ and target $Y$. Then for each interactive training step, we have for both agents

\begin{equation}
\mathcal{L}_{sender}^{\S2P} = \mathcal{L}^{\GSTE}(\text{Fr}^{ft}, \text{De}^{ft}) + \alpha \mathcal{L}^{\SU}(\text{Fr}^{pre}, \text{En}^{pre}) 
\end{equation}
\begin{equation}
\mathcal{L}_{receiver}^{\S2P} = \mathcal{L}^{\GSTE}(\text{Fr}^{ft}, \text{De}^{ft}) + \alpha \mathcal{L}^{\SU}(\text{En}^{pre}, \text{De}^{pre}) 
\end{equation}

\begin{figure}[ht]
    \centering
  \begin{subfigure}{0.4\columnwidth}
    \includegraphics[width=\textwidth]{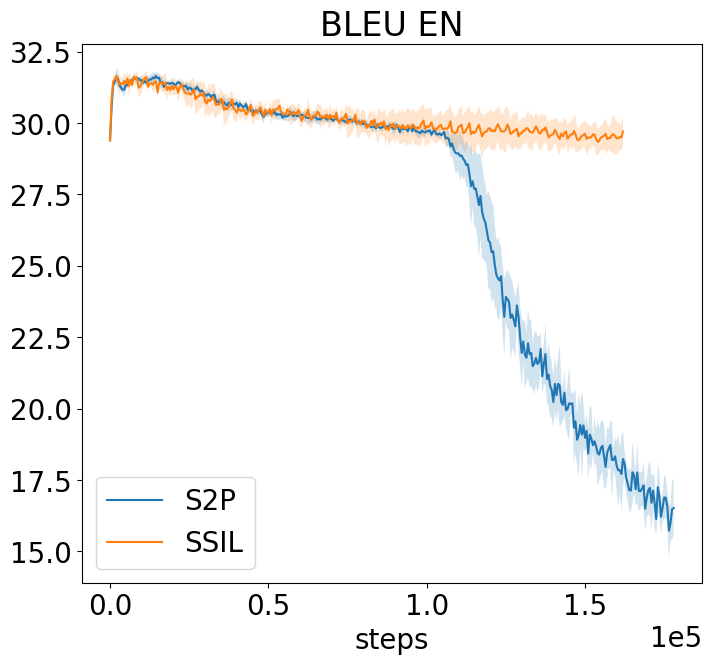}
    \caption{Bleu En}
  \end{subfigure}
  \hspace{0.1\columnwidth}
  \begin{subfigure}{0.4\columnwidth}
    \includegraphics[width=\textwidth]{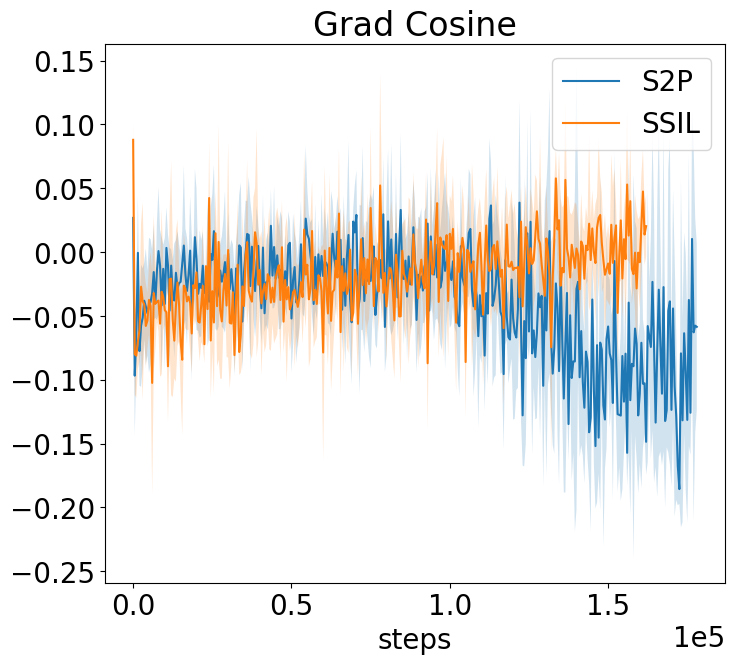}
    \caption{Cosine Similarity}
  \end{subfigure}
  \caption{Cosine similarity bewteen $\mathcal{L}^{\SU}_{pretrain}$ and $\mathcal{L}^{\INT}$ when $\alpha=0.7$} 
  \label{fig:grad_cosine_0_7}
 \end{figure}

\section{Translation Game Implementation Details}
\label{sec:translation_game}
We here report the experimenatl protocol from
We use the Moses tokenizer~\cite{koehn-etal-2007-moses} and we learn a byte-pair-encoding~\cite{sennrich-etal-2016-neural} from Multi30K with all language. Then the same BPE is applied to different dataset. Our vocab size for En, Fr, De is 11552, 13331, and 12124. Our pretraining datasets are IWSLT while the finetuning datasets are Multi30K. Our language model is trained with captions data from MSCOCO~\cite{lin2014microsoft}. For image ranker, we use the captions in Multi30K as well as the original Flickr30K images. We use a ResNet152 with pretrained ImageNet weights to extract the image features. We also normalize the image features. We follow the pretraining and model architecture from work~\cite{lu2020countering}.

\section{Hyper-parameters}
During finetuning, we set Gumbel temperature to be 0.5 and follow the previous work~\cite{lu2020countering} for other hyperparameters, e.g. learning rate, batch size, etc. We list our hyper-parameters and our sweep:
\begin{table}[h]
\centering
\begin{tabular}{ll}
\toprule
Name  & Sweep                                      \\ \midrule
$k_1$    & 3000, 4000                                 \\ 
$k_2$    & 200, 300, 400                              \\
$k_2'$   & 200, 300, 400                              \\
$\alpha$ & 0, 0.01, 0.1, 0.2, 0.3, 0.4, 0.5, 0.6, 0.7 \\ \bottomrule
\end{tabular}
\end{table}
We mainly use P100 GPU for our experiments. For training 200k steps, Gumbel takes 17 hours, S2P takes 24 hours, SIL takes 18 hours and \algo takes 24 hours.
The best hyperparameters for SIL are $k_1=3000, k_2=200, k_2'=300$. The best $alpha$ for $S2P$ is $1$, while for $SSIL$ we choose $\alpha=0.5$.
\section{S2P Details}
\label{sec:s2p}
\begin{figure}[ht]
\centering
  \begin{subfigure}{0.4\columnwidth}
  \includegraphics[width=\textwidth]{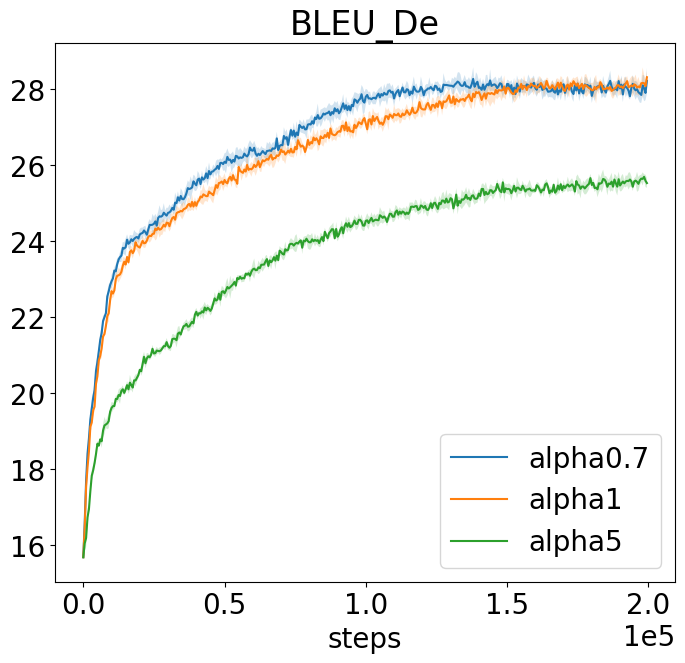}
  \caption{BLEU De (Task Score)}
  \end{subfigure}
  \hspace{0.1\columnwidth}
  \begin{subfigure}{0.4\columnwidth}
  \includegraphics[width=\textwidth]{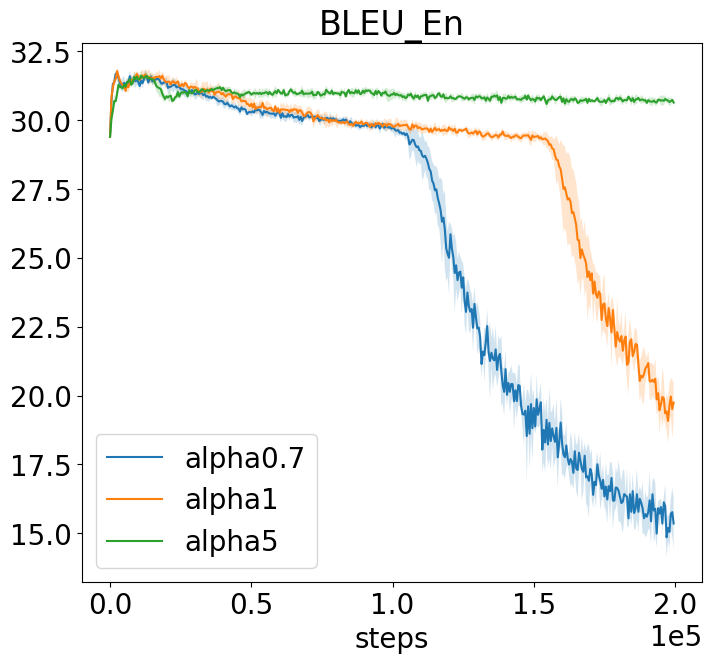}
  \caption{BLEU En} 
  \end{subfigure} 

  \caption{S2P with different $\alpha$. Increased $\alpha$ might delay or remove the late-stage collapse, but it might be at the cost of task score.}
  \label{fig:s2p}
  \end{figure}

We show the results of S2P with varying $\alpha$ in Figure~\ref{fig:s2p}. In general, one can find that for S2P there is a trade-off between grounding score and task score controlled by $\alpha$. A larger $\alpha$ might delay the eventual collapse. However, if the $\alpha$ is too large, the task score will decrease significantly. As a result, even though increasing $\alpha$ seems to fit the intuition, it cannot fix the problem. 

\begin{figure}[ht]
    \centering
  \begin{subfigure}{0.24\columnwidth}
  \includegraphics[width=\textwidth]{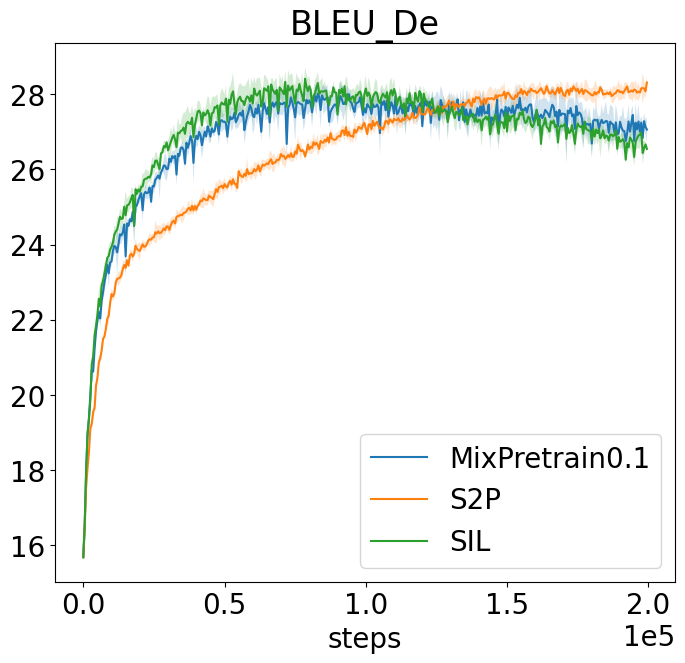}
  \caption{BLEU De (Task Score)}
  \end{subfigure}
  \hfill
  \begin{subfigure}{0.24\columnwidth}
  \includegraphics[width=\textwidth]{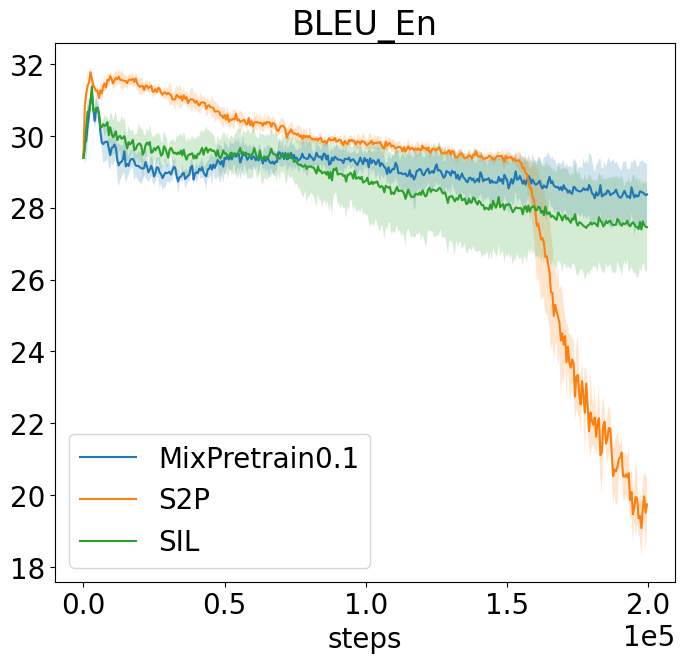}
  \caption{BLEU En} 
  \end{subfigure} 
  \hfill
  \begin{subfigure}{0.24\columnwidth} 
  \includegraphics[width=\textwidth]{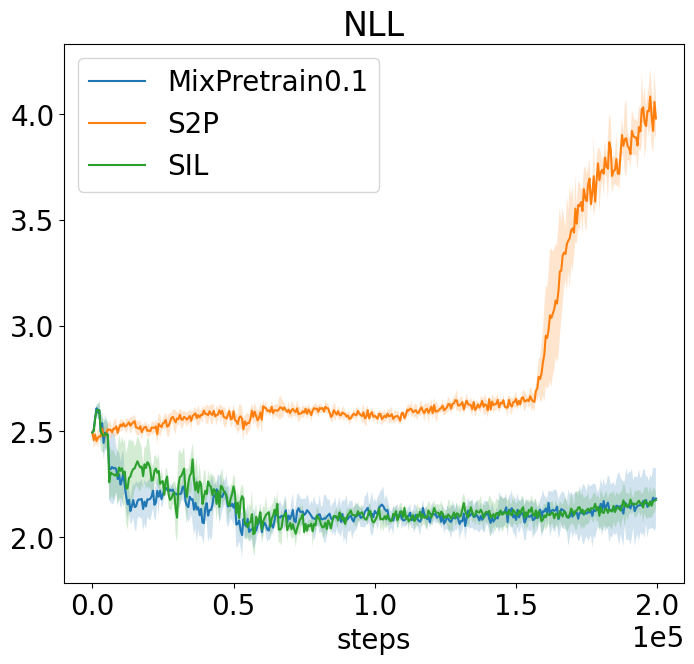} 
  \caption{NLL} 
  \end{subfigure}  
  \hfill 
  \begin{subfigure}{0.24\columnwidth} 
  \includegraphics[width=\textwidth]{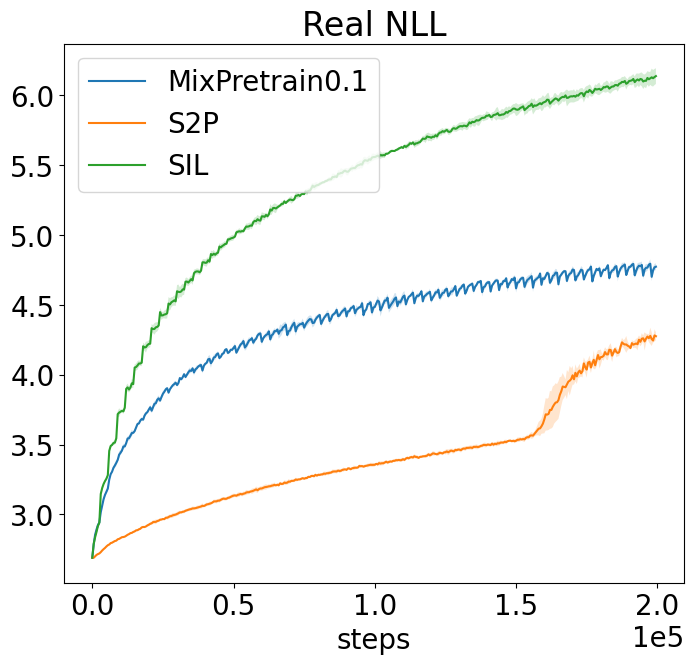} 
  \caption{RealNLL} 
  \end{subfigure}
  \caption{Mix with Pretraining data in SIL.}
  \label{fig:mixdata}
\end{figure}

 \begin{figure}[h!]
  \centering
  \begin{subfigure}{0.24\columnwidth}
  \includegraphics[width=\textwidth]{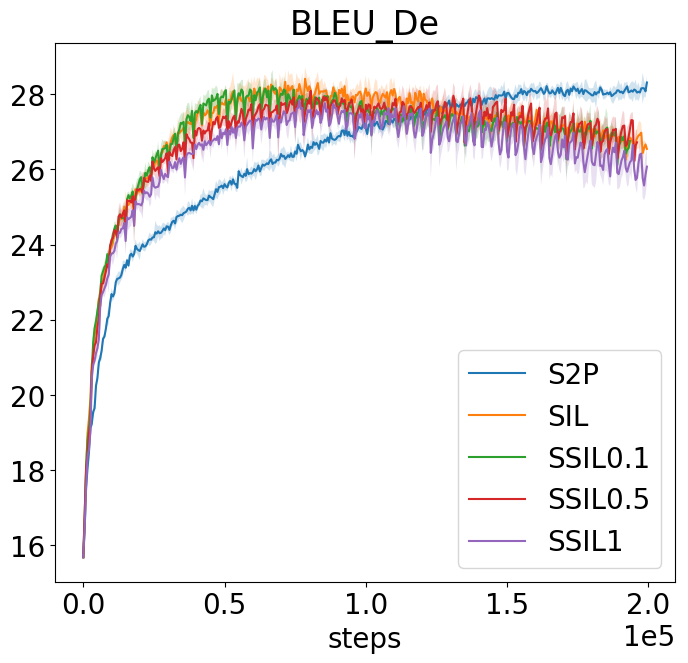}
  \caption{BLEU De (Task Score)}
  \end{subfigure}
  \hfill
  \begin{subfigure}{0.24\columnwidth}
  \includegraphics[width=\textwidth]{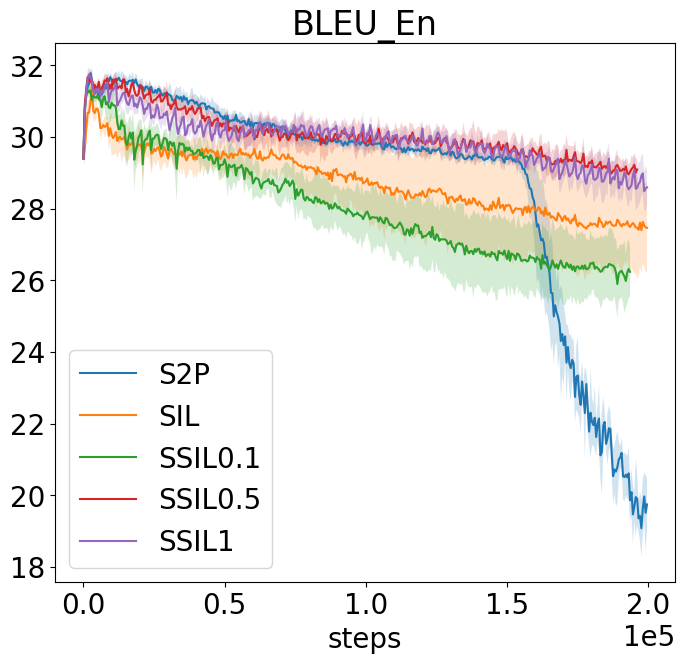}
  \caption{BLEU En} 
  \end{subfigure} 
  \hfill
  \begin{subfigure}{0.24\columnwidth} 
  \includegraphics[width=\textwidth]{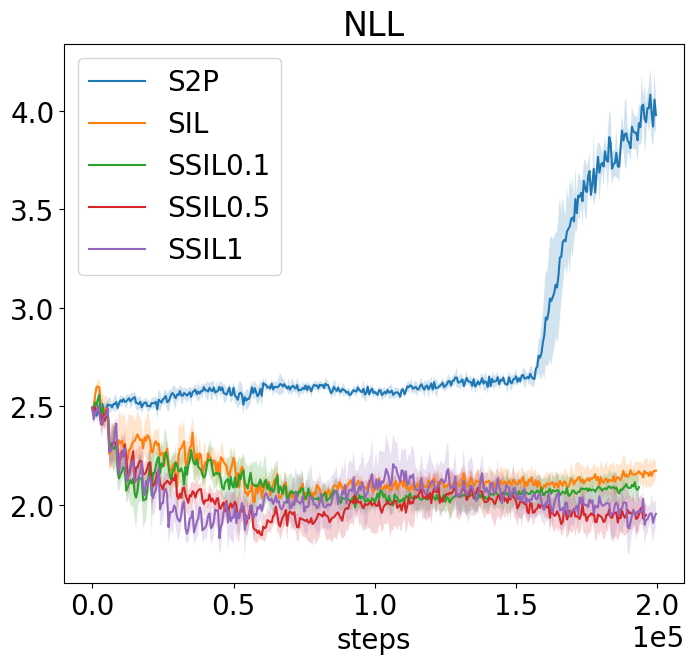} 
  \caption{NLL} 
  \end{subfigure}  
  \hfill 
  \begin{subfigure}{0.24\columnwidth} 
  \includegraphics[width=\textwidth]{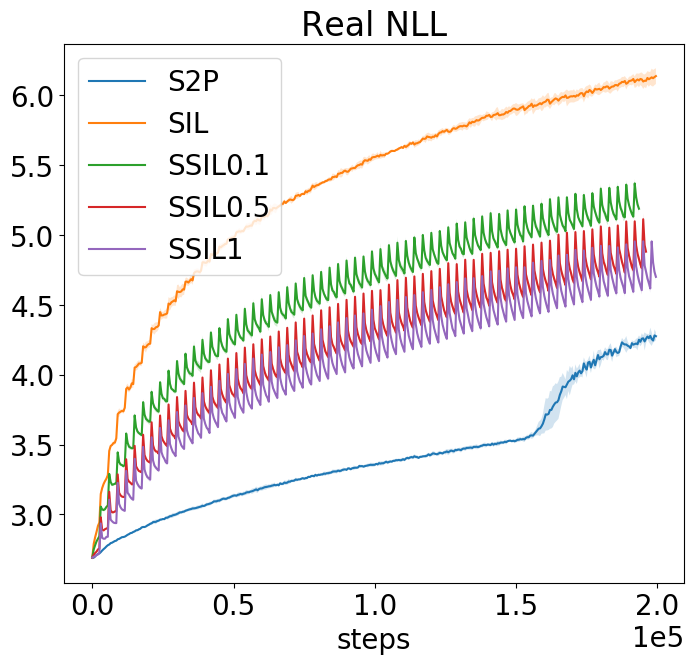} 
  \caption{RealNLL} 
  \end{subfigure}
  \caption{\algo with different $\alpha$}
\end{figure}

\begin{figure}[ht]
    \centering
  \begin{subfigure}{0.35\columnwidth}
      \includegraphics[width=\textwidth]{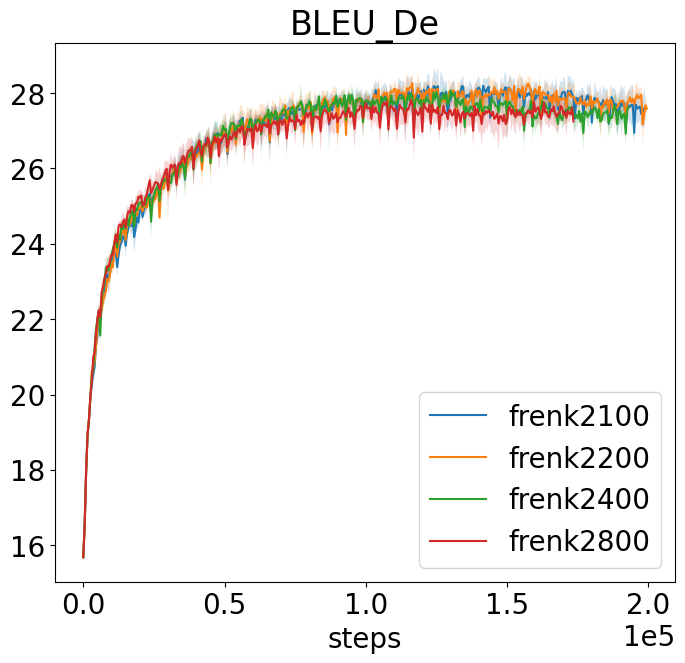}
      \caption{BLEU De (Task Score)}
      \end{subfigure}
   \begin{subfigure}{0.35\columnwidth}
      \includegraphics[width=\textwidth]{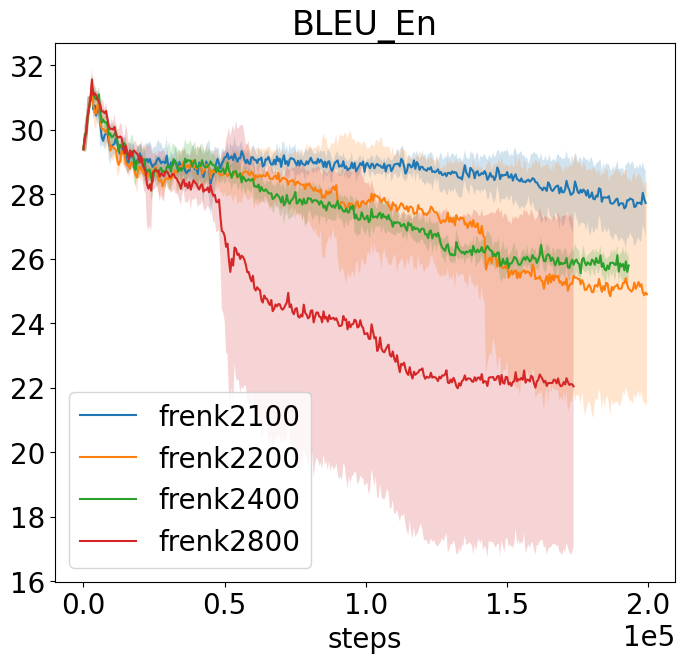}
      \caption{BLEU En} 
  \end{subfigure} 
  \caption{Effect of $k_2$ for MixData.$\alpha=0.2$ }
\end{figure}

\begin{figure}[ht]
    \centering
  \begin{subfigure}{0.35\columnwidth}
      \includegraphics[width=\textwidth]{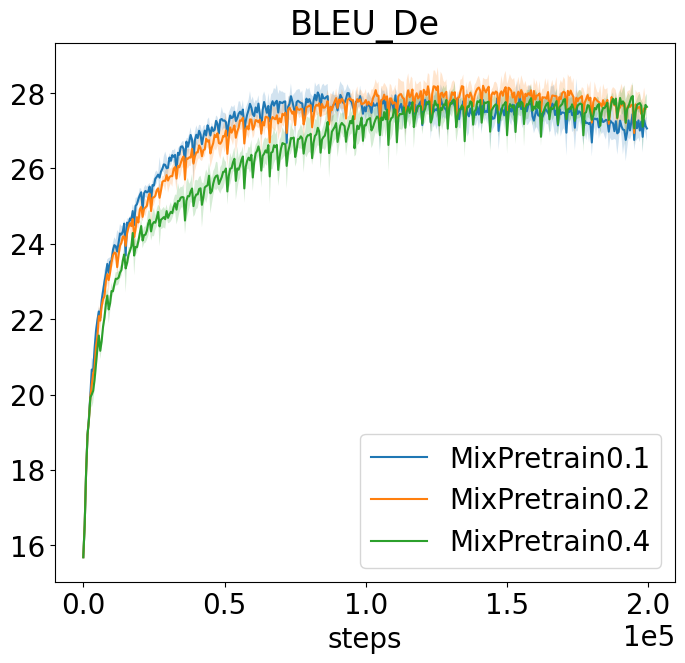}
      \caption{BLEU De (Task Score)}
  \end{subfigure}
      \begin{subfigure}{0.35\columnwidth}
      \includegraphics[width=\textwidth]{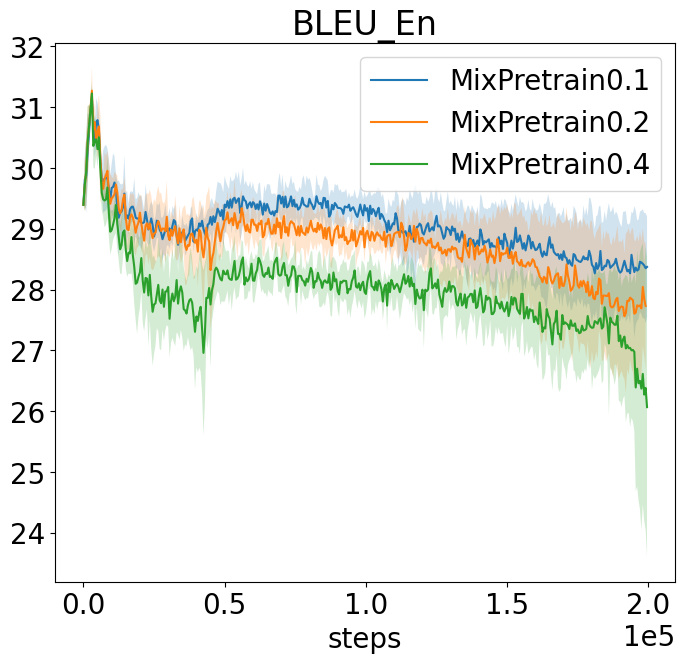}
      \caption{BLEU En} 
  \end{subfigure} 
  \caption{Effect of $\alpha$ for MixData. $k_2=100$}
  \label{fig:mixdata_alpha}
\end{figure}

\end{document}